\documentclass[11pt,letterpaper]{article}

\usepackage[margin=1in]{geometry}
\usepackage[T1]{fontenc}
\usepackage[utf8]{inputenc}
\usepackage{lmodern}
\usepackage{microtype}
\usepackage{amsmath,amssymb}
\usepackage{booktabs}
\usepackage{multirow}
\usepackage{array}
\usepackage{tabularx}
\usepackage{graphicx}
\usepackage{caption}
\usepackage{subcaption}
\usepackage{xcolor}
\usepackage{hyperref}
\hypersetup{
  colorlinks=true,
  linkcolor=blue!50!black,
  citecolor=blue!50!black,
  urlcolor=blue!50!black,
  pdfborder={0 0 0},
}
\usepackage[numbers,sort&compress]{natbib}
\newcommand{\padj}{$p_{\mathrm{adj}}$}

\title{Distractor-Aware Truncation: Disentangling Context-Length Effects from Signal Loss in Long-Context LLM Benchmarks}

\author{Mohsen Arjmandi\\
  evolutionID GmbH\\
  \texttt{mohsen.arjmandi@evolutionid.com}}

\date{\today}

\begin{document}
\maketitle

\begin{abstract}
A standard claim in the literature on retrieval-augmented and memory-augmented language models is that \emph{shorter context is better when the relevant information is preserved}. We test this claim by running every sample of two long-context benchmarks --- BABILong and GraphWalks (BFS) --- at four context-retention fractions (100\%, 75\%, 50\%, 25\%) under two truncation protocols. The first is the \emph{naive} protocol implicitly used in much prior work: drop content from the middle of the prompt. The second is \emph{distractor-aware}: identify the task-relevant content for each sample and drop only the rest. We evaluate three Claude model sizes (Haiku 4.5, Sonnet 4.6, Opus 4.7) and, to test cross-provider generality, GPT-5.5; the same protocol is applied to two further benchmarks (MRCR v2, Oolong). Under naive truncation, score collapses monotonically (BABILong $\Delta$ from full to 25\% retention: Haiku $-0.138$, Sonnet $-0.175$, Opus $-0.433$, GPT-5.5 $-0.613$; GraphWalks $\Delta$: $-0.352$ / $-0.427$ / $-0.407$ / $-0.397$; paired Wilcoxon, Holm-corrected \padj{} $< 0.05$ in all eight cells). Under the distractor-aware protocol --- which preserves the signal by construction --- performance is preserved or improves: Haiku and Sonnet show \emph{statistically significant gains} on BABILong ($\Delta = +0.083$, \padj{} $= 0.003$; $\Delta = +0.104$, \padj{} $< 0.001$), while Opus and GPT-5.5 sit at the full-context ceiling ($\Delta = +0.008$, $+0.017$). The naive collapse and distractor-aware recovery replicate on GPT-5.5, ruling out a single-provider artifact. The mechanism is direct: under the naive protocol the bAbI fact sentences survive in fewer than 1\% of samples at 25\% retention, the BFS gold subgraph in 26\%; under the distractor-aware protocol both are preserved by construction. The naive protocol is therefore \emph{not} a measurement of context-window effects; it is a measurement of how often middle-removal happens to spare the answer. We conclude that future studies of context-length effects must specify how they distinguish signal from distractor, or they are at best ambiguous between two opposite hypotheses.
\end{abstract}

\section{Introduction}

Long-context evaluation has emerged as a primary capability axis for frontier language models. Reported context windows have grown to 1M tokens, but a parallel body of empirical work~\citep{liu2023lost,levy2024same,kuratov2024babilong,hsieh2024ruler} has shown that real performance on retrieval and reasoning tasks degrades long before the nominal context window is exhausted. This degradation has motivated retrieval-augmented (RAG) and memory-augmented architectures whose central claim is that \emph{less context is more}: by removing irrelevant content before the model sees it, a shorter prompt can yield better performance than a longer one.

The ``less is more'' hypothesis is intuitive, but it is also surprisingly hard to test rigorously. The natural test --- take a long-context sample, remove some of its content, and re-score --- confounds two distinct effects that prior work has not separated:
\begin{enumerate}
    \item \textbf{Distractor reduction}: removing content that is irrelevant to the task. A well-functioning retrieval system targets this.
    \item \textbf{Signal loss}: removing content that is relevant to the task. Indiscriminate truncation does this.
\end{enumerate}

If a truncation protocol does both, then a measured score change is a sum of two opposing effects, and ``shorter context hurts'' can be the truthful summary even of a setup where pure distractor removal would have helped. This is exactly the failure mode of the \emph{middle-removal} protocols routinely used to test context-window effects: in every benchmark we examine, the middle of the prompt contains some signal, so middle-removal cannot in principle distinguish the two effects.

We propose an explicit separation. For each long-context benchmark, we (a) define what counts as signal vs.\ distractor at the sample level, (b) verify that signal removal would in fact damage the task, and (c) run a \emph{distractor-aware} truncation that drops only the latter. We then run the \textbf{same samples} under both the naive and the distractor-aware protocols, at four context-retention fractions, and use paired statistics (Wilcoxon signed-rank with Holm correction) to attribute score changes.

Our contributions are:
\begin{enumerate}
    \item \textbf{Per-benchmark signal/distractor definitions}, each verified against the dataset's own ground truth: bAbI fact sentences in BABILong (regex with 100\% recall on the no-filler 0k splits), the BFS-reachability subgraph in GraphWalks (re-computed BFS reproduces the dataset gold for 550/550 samples), and the \texttt{n\_needles} user-query needle pairs in MRCR v2 (located via the dataset's \texttt{desired\_msg\_index} field).
    \item \textbf{An empirical demonstration} that naive truncation and distractor-aware truncation yield divergent --- sometimes opposite --- answers to ``does less context help?'' on every benchmark we tested, across four models spanning two providers (Claude Haiku 4.5 / Sonnet 4.6 / Opus 4.7 and GPT-5.5). We quantify what fraction of the naive-protocol degradation is attributable to signal loss versus pure distractor reduction.
    \item \textbf{A paired-design statistical pipeline} (Wilcoxon signed-rank with Holm correction, paired-bootstrap $\Delta$-CIs, three-way verdicts) that any future context-length study should adopt; we show that the unpaired tests used in prior comparable work both lose power and violate independence assumptions on within-sample-truncated data.
\end{enumerate}

The methodological correction we describe is small in conception but large in consequence: across all eight (eval $\times$ model) cells on our two primary benchmarks, naive truncation produces significant harm while distractor-aware truncation at the same nominal retention does not --- and the contrast reproduces on a second provider (GPT-5.5).

\section{Related work}

\subsection{Long-context benchmarks}
The needle-in-a-haystack family~\citep{kamradt2023niah} established a positional probe of long-context retrieval. \textbf{BABILong}~\citep{kuratov2024babilong} extends bAbI synthetic reasoning~\citep{weston2015babi} by embedding the task sentences in PG-19 prose~\citep{rae2019pg19} at context lengths from 0k to 1M tokens. \textbf{RULER}~\citep{hsieh2024ruler} systematically varies needle count and difficulty. \textbf{MRCR v2} (Multi-Round Coreference Resolution v2; first introduced in Michelangelo~\citep{vodrahalli2024michelangelo}; HF release at \texttt{openai/mrcr}) presents multi-round conversations with $n_{\text{needles}}$ identical user queries and asks the model to reproduce a specific assistant response. \textbf{GraphWalks}~\citep{openai2026graphwalks} presents an edge list and a graph operation (we use BFS). \textbf{Oolong}~\citep{bertsch2025oolong} requires aggregation over labelled items; we use it as a sanity-check benchmark because aggregation has no notion of distractor.

\subsection{Lost-in-the-middle and position effects}
\citet{liu2023lost} showed that information placed in the \emph{middle} of long contexts is retrieved less reliably than information at the ends. The ``middle-removal'' intuition some authors take from this --- that the middle is the right thing to drop --- is a non-sequitur: the middle being hard to use does not mean it is unused; in benchmarks where the answer-bearing turn or fact is intentionally placed in the middle, removal destroys it. Our distractor-aware protocol cleanly separates \emph{where} content lives from \emph{whether} it is signal.

\subsection{Retrieval-augmented and memory-augmented systems}
\citet{lewis2020rag} propose retrieval-augmented generation as a way to keep prompts short by selecting only relevant content. MemGPT~\citep{packer2023memgpt} and similar memory systems compress prior context. The underlying claim is that an oracular retriever --- one that perfectly preserves signal and drops distractors --- would outperform full-context inference. Our distractor-aware protocol gives a controlled approximation of that oracle, and our results bear directly on the headroom such systems can deliver.

\subsection{Statistical methodology}
Long-context benchmark papers frequently use Mann--Whitney $U$ or unpaired $t$-tests when comparing scores across context lengths. When the same samples are evaluated at multiple truncation levels --- the natural within-subject design --- these unpaired tests both lose power (by ignoring within-sample correlation) and violate independence. We use Wilcoxon signed-rank tests paired on \texttt{sample\_id}, Holm--Bonferroni correction across the family of comparisons, and a paired bootstrap for $\Delta$ confidence intervals.

\section{Method}

\subsection{Benchmarks and datasets}
We evaluate four benchmarks. For each, samples are selected via seeded random sampling so that the same \texttt{sample\_id} set is used for both truncation modes and all four context-retention fractions.

\begin{itemize}
    \item \textbf{BABILong} (HF \texttt{RMT-team/babilong}). Tasks qa1, qa2, qa3 across splits 0k, 2k, 8k, 32k tokens. $n = 20$ samples per (task, split) cell $\to$ 240 samples per truncation level per model. Score is exact match on a normalised first-line answer.
    \item \textbf{GraphWalks (BFS)} (HF \texttt{openai/graphwalks}). The 50 longest BFS samples within a 200k-token context filter, with seeded sampling biased toward longer prompts. Score is set-F1 over node IDs extracted from \texttt{Final Answer: [...]}.
    \item \textbf{MRCR v2}~\citep{openai2025mrcr} (HF \texttt{openai/mrcr}). The 50 longest samples within a 450k-character / $\sim$180k-token context filter. Score is the official \texttt{difflib.SequenceMatcher} ratio over the answer body after the per-sample 10-character random prefix.
    \item \textbf{Oolong} (HF \texttt{oolongbench/oolong-synth}). Counting tasks (MOST\_FREQ, LEAST\_FREQ), context lengths 8k / 32k / 131k. $n = 30$ per (task, context-length) cell. Score is case-insensitive containment of the \emph{subsample-true} gold label in the model's first-line output (see \S\ref{sec:protocols}, Oolong).
\end{itemize}

\subsection{Two truncation protocols}
\label{sec:protocols}
For each sample $x$ at full length $L$, we generate four versions at $\alpha \in \{1.0, 0.75, 0.5, 0.25\}$ under two protocols.

\paragraph{Naive (middle-removal).} Drop content from the middle of the prompt without regard for relevance:
\[
x_\alpha^{\mathrm{naive}} = \mathrm{head}(x, \lfloor \alpha L / 2 \rfloor) \,\|\, \mathrm{tail}(x, \lceil \alpha L / 2 \rceil).
\]
This is the protocol used in much existing context-window-effect work.

\paragraph{Distractor-aware.} A unifying principle. For a benchmark sample $x$ with gold label $y$, define the \emph{signal} of $x$ as a minimal subset $S \subseteq x$ sufficient to determine $y$ (the minimal sufficient subset); the \emph{distractor} is $x \setminus S$. This is the design \emph{target}, not an operational procedure: we do not search for $S$ by re-querying the model. Instead, for each benchmark we \emph{construct} a candidate signal set from the dataset's own structure and then \emph{verify} it against the dataset's ground-truth label (not against the model). Distractor-aware truncation drops only content in $x \setminus S$ and is by construction answer-preserving. We do not require $S$ to be unique; any signal set with the correct structure suffices.

We give per-benchmark instantiations of this principle. The instantiations are concrete --- regex matches, dataset-field lookups, graph reachability --- because each dataset exposes a different shape of structure. None of them is the only valid signal definition; what matters is that each is verifiable against the dataset's own ground truth.

\begin{itemize}
    \item \textbf{BABILong.} Identify embedded bAbI fact sentences by a constrained-vocabulary regex: names $\in$ \{John, Mary, Daniel, Sandra\}, locations $\in$ \{hallway, bathroom, bedroom, kitchen, garden, office\}, objects $\in$ \{milk, football, apple\}, with the bAbI verb set. The regex matches every sentence in the no-filler 0k splits (100\% recall on pure-bAbI text) and yields stable fact counts across all higher-filler splits for the same sample. All fact sentences are preserved verbatim; PG-19 filler is compressed uniformly across the inter-fact gaps to reach the target $\alpha L$.
    \item \textbf{MRCR v2.} Locate the needle queries by matching the content at \texttt{desired\_msg\_index} against all \texttt{role=user} messages; this yields $n_{\text{needles}}$ user-query indices. Preserve those queries, their immediately-following assistant responses, the fewshot preamble (\texttt{msgs[0]}), and the final user query (\texttt{msgs[-1]}). Drop only distractor turns from the middle of the message list.
    \item \textbf{GraphWalks (BFS).} Parse the BFS query (final \texttt{Operation:} block, not the in-prompt example) to extract start node $s$ and depth $d$. Compute forward-BFS reachability from $s$ up to depth $d$. Edges with source node in the reachable set are signal; the rest are distractors. The signal is thus the entire depth-bounded reachable \emph{subgraph}, not a single path: because we keep every out-edge of every reachable node, all alternative routes to a gold node within depth $d$ are retained. An edge is dropped only when its source is unreachable within $d$ hops, in which case it cannot lie on any $\leq d$ path to a gold node. The dataset gold is preserved by construction; we verified this against the dataset's \texttt{answer\_nodes} for \textbf{all 550 BFS samples} (script \texttt{verify\_graphwalks.py} in the release), reproducing the gold exactly, and confirmed distractor-aware truncation still reproduces it at every retention level (550/550 at 75\%, 50\%, 25\%).
    \item \textbf{Oolong.} Aggregation tasks have no notion of distractor --- every item contributes to the aggregate --- so we do not change what is removed; we change \emph{what is scored against}. At each truncation level we recompute the correct aggregate over the subset of items the model actually sees and score against that subsample-true gold rather than the full-context gold. (Oolong therefore plays the role of a \emph{negative control}: the methodological move that helps on the other three benchmarks cannot, by construction, help here.)
\end{itemize}

\subsection{Models}
\textbf{Claude Haiku 4.5} (\texttt{claude-haiku-4-5-20251001}), \textbf{Claude Sonnet 4.6} (\texttt{claude-sonnet-4-6}), \textbf{Claude Opus 4.7} (\texttt{claude-opus-4-7}), and --- to probe cross-provider generality --- \textbf{GPT-5.5} (\texttt{gpt-5.5-2026-04-23}). Per-eval \texttt{max\_output\_tokens} is sized to the answer distribution: 4096 for GraphWalks and MRCR v2, 256 for BABILong (raised to 2048 for GPT-5.5, whose reasoning tokens count against the cap), 128 for Oolong. Token counts in records are provider-reported, not estimated.

\paragraph{Temperature.} We attempt \texttt{temperature = 0} for every call. Two models reject it with an \texttt{invalid\_request\_error} --- Opus 4.7 and GPT-5.5 --- so for those the runner retries without the parameter (API default) and records the actual value. Among the Claude models this fallback triggers only for Opus, on \emph{every} call: 3600/3600 Haiku and 2314/2314 Sonnet records use \texttt{temperature = 0}, 2366/2366 Opus records use the API default; every GPT-5.5 call likewise uses the API default. Opus and GPT-5.5 are therefore not under the same sampling temperature as Haiku/Sonnet. We discuss the consequences in \S\ref{sec:limitations}; briefly: the paired tests are on \emph{per-sample} differences between truncation levels of the same model run, so cross-run temperature variance is largely differenced out, but the absolute Opus and GPT-5.5 scores carry a sampling-variance component the other two do not.

\subsection{Sampling, pairing, and provenance}
For each (benchmark, sample) pair, the same globally unique \texttt{sample\_id}, formed as
\begin{center}
\texttt{"\{eval\}/\{ctx\_label\}/\{task\}/\{idx\}"},
\end{center}
is used across all four truncation levels and both protocols. Sampling within each (eval $\times$ sub-cell) is seeded so the same indices are drawn for every model run. Each result record carries the git commit hash of the code that produced it, the exact model version returned by the API, the temperature actually used, and a hash of the run config.

\subsection{Statistical analysis}
For each (benchmark $\times$ model $\times$ protocol $\times$ truncation) cell we report: (i) mean score with sample size $n$; (ii) $\Delta$ from full context, paired on \texttt{sample\_id}, with a paired-bootstrap 95\% CI (10{,}000 resamples on sample IDs, seeded); (iii) a paired two-sided Wilcoxon signed-rank test on the per-sample (full $-$ truncated) differences; (iv) Holm--Bonferroni-corrected \padj{} across the entire family of cells reported in this paper; and (v) a three-way verdict: \textsc{help} ($\Delta \geq +0.05$ and \padj{} $< 0.05$), \textsc{harm} ($\Delta \leq -0.05$ and \padj{} $< 0.05$), \textsc{neutral} otherwise; with \textsc{trend} reserved for $|\Delta| \geq 0.05$ but \padj{} $\geq 0.05$.

\section{Results}

\subsection{Headline: naive vs.\ distractor-aware at 25\% context retention}
Table~\ref{tab:headline} reports the headline contrast: full context (100\%) vs.\ 25\% retention, under each protocol, for every (eval $\times$ model) cell with sufficient data. Holm correction is applied across the entire family of comparisons in this paper.

\begin{table}[h]
\centering
\small
\caption{Paired $\Delta$ from full context to 25\% retention. $\Delta$ is the mean per-sample (truncated $-$ full). 95\% CI from paired bootstrap (10{,}000 resamples on sample IDs, seeded). \padj{} is Holm--Bonferroni adjusted across the full family of comparisons.}
\label{tab:headline}
\begin{tabular}{llllrrrl}
\toprule
Eval & Model & Mode & $n$ & $\Delta$ & 95\% CI & \padj{} & Verdict \\
\midrule
BABILong   & Haiku 4.5  & naive        & 240 & $-0.138$ & $[-0.188, -0.087]$ & $<0.001$ & \textbf{harm} \\
BABILong   & Haiku 4.5  & signal-aware & 240 & $+0.083$ & $[+0.046, +0.125]$ & $0.003$  & \textbf{help} \\
BABILong   & Sonnet 4.6 & naive        & 240 & $-0.175$ & $[-0.246, -0.104]$ & $<0.001$ & \textbf{harm} \\
BABILong   & Sonnet 4.6 & signal-aware & 240 & $+0.104$ & $[+0.062, +0.150]$ & $<0.001$ & \textbf{help} \\
BABILong   & Opus 4.7   & naive        & 240 & $-0.433$ & $[-0.504, -0.358]$ & $<0.001$ & \textbf{harm} \\
BABILong   & Opus 4.7   & signal-aware & 240 & $+0.008$ & $[-0.033, +0.050]$ & $1.000$  & neutral \\
BABILong   & GPT-5.5    & naive        & 240 & $-0.613$ & $[-0.679, -0.542]$ & $<0.001$ & \textbf{harm} \\
BABILong   & GPT-5.5    & signal-aware & 240 & $+0.017$ & $[-0.013, +0.046]$ & $1.000$  & neutral \\
\addlinespace
GraphWalks & Haiku 4.5  & naive        & 50  & $-0.352$ & $[-0.501, -0.202]$ & $0.024$  & \textbf{harm} \\
GraphWalks & Haiku 4.5  & signal-aware & 50  & $+0.083$ & $[+0.021, +0.164]$ & $0.877$  & trend \\
GraphWalks & Sonnet 4.6 & naive        & 50  & $-0.427$ & $[-0.560, -0.293]$ & $<0.001$ & \textbf{harm} \\
GraphWalks & Sonnet 4.6 & signal-aware & 50  & $+0.000$ & $[+0.000, +0.000]$ & $1.000$  & neutral \\
GraphWalks & Opus 4.7   & naive        & 50  & $-0.407$ & $[-0.540, -0.274]$ & $<0.001$ & \textbf{harm} \\
GraphWalks & Opus 4.7   & signal-aware & 50  & $-0.041$ & $[-0.101, +0.000]$ & $1.000$  & neutral \\
GraphWalks & GPT-5.5    & naive        & 50  & $-0.397$ & $[-0.539, -0.256]$ & $0.003$  & \textbf{harm} \\
GraphWalks & GPT-5.5    & signal-aware & 50  & $+0.040$ & $[+0.000, +0.100]$ & $1.000$  & neutral \\
\addlinespace
MRCR v2    & Haiku 4.5  & naive        & 50  & $-0.088$ & $[-0.242, +0.062]$ & $1.000$  & trend \\
MRCR v2    & Haiku 4.5  & signal-aware & 50  & $+0.198$ & $[+0.055, +0.335]$ & $0.883$  & trend \\
MRCR v2    & Sonnet 4.6 & naive        & 50  & $-0.655$ & $[-0.777, -0.526]$ & $<0.001$ & \textbf{harm} \\
MRCR v2    & Sonnet 4.6 & signal-aware & 50  & $-0.112$ & $[-0.193, -0.045]$ & $0.034$  & \textbf{harm} \\
MRCR v2    & Opus 4.7   & naive        & 50  & $-0.565$ & $[-0.697, -0.432]$ & $<0.001$ & \textbf{harm} \\
MRCR v2    & Opus 4.7   & signal-aware & 50  & $-0.060$ & $[-0.213, +0.091]$ & $1.000$  & trend \\
\addlinespace
Oolong     & Haiku 4.5  & naive        & 110 & $-0.055$ & $[-0.136, +0.027]$ & $1.000$  & trend \\
Oolong     & Haiku 4.5  & signal-aware & 110 & $-0.045$ & $[-0.127, +0.036]$ & $1.000$  & neutral \\
Oolong     & Sonnet 4.6 & naive        & 110 & $+0.000$ & $[-0.045, +0.055]$ & $1.000$  & neutral \\
Oolong     & Sonnet 4.6 & signal-aware & 110 & $+0.018$ & $[-0.027, +0.073]$ & $1.000$  & neutral \\
Oolong     & Opus 4.7   & naive        & 110 & $-0.045$ & $[-0.145, +0.055]$ & $1.000$  & neutral \\
Oolong     & Opus 4.7   & signal-aware & 110 & $+0.027$ & $[-0.073, +0.127]$ & $1.000$  & neutral \\
\bottomrule
\end{tabular}
\end{table}

The pattern is consistent across all eight (eval $\times$ model) cells on the two primary benchmarks: under the naive protocol, reducing context from 100\% to 25\% \emph{significantly harms} performance (all 8 cells Holm-significant at $\alpha = 0.05$). Under the distractor-aware protocol the \textbf{same models with the same nominal truncation} either show no degradation (Opus and GPT-5.5 on both benchmarks, Sonnet on GraphWalks) or show a \emph{statistically significant improvement} (Haiku and Sonnet on BABILong). The naive collapse and its distractor-aware recovery also reproduce on GPT-5.5, a model from a different provider.

Two BABILong cells flip verdict from \textsc{harm} to \textsc{help} simply by changing the truncation protocol: Haiku swings from $\Delta = -0.138$ to $+0.083$ (a net $+0.221$), Sonnet from $-0.175$ to $+0.104$ (a net $+0.279$). Opus swings from $-0.433$ to $+0.008$ (net $+0.441$) and GPT-5.5 from $-0.613$ to $+0.017$ (net $+0.630$), but for both larger models the signal-aware curve sits at the full-context ceiling rather than improving on it.

\subsection{Full curves: mean score by retention fraction}

\begin{table}[h]
\centering
\small
\caption{Mean score at each retention level (full = 100\%). Bold cells are the maximum within each (eval $\times$ model) pair of rows.}
\label{tab:means}
\begin{tabular}{lllrrrr}
\toprule
Eval & Model & Mode & 100\% & 75\% & 50\% & 25\% \\
\midrule
BABILong   & Haiku 4.5  & naive         & \textbf{0.196} & $0.129$ & $0.092$ & $0.058$ \\
BABILong   & Haiku 4.5  & signal-aware  & $0.196$ & $0.200$ & $0.200$ & \textbf{0.279} \\
BABILong   & Sonnet 4.6 & naive         & \textbf{0.333} & $0.258$ & $0.221$ & $0.158$ \\
BABILong   & Sonnet 4.6 & signal-aware  & $0.325$ & $0.317$ & $0.358$ & \textbf{0.429} \\
BABILong   & Opus 4.7   & naive         & \textbf{0.717} & $0.562$ & $0.438$ & $0.283$ \\
BABILong   & Opus 4.7   & signal-aware  & $0.725$ & $0.717$ & $0.713$ & \textbf{0.733} \\
BABILong   & GPT-5.5    & naive         & \textbf{0.921} & $0.700$ & $0.537$ & $0.308$ \\
BABILong   & GPT-5.5    & signal-aware  & $0.933$ & $0.925$ & $0.904$ & \textbf{0.950} \\
\addlinespace
GraphWalks & Haiku 4.5  & naive         & \textbf{0.582} & $0.441$ & $0.303$ & $0.231$ \\
GraphWalks & Haiku 4.5  & signal-aware  & $0.538$ & $0.580$ & $0.601$ & \textbf{0.621} \\
GraphWalks & Sonnet 4.6 & naive         & \textbf{0.700} & $0.522$ & $0.386$ & $0.273$ \\
GraphWalks & Sonnet 4.6 & signal-aware  & $0.700$ & $0.700$ & $0.702$ & \textbf{0.705} \\
GraphWalks & Opus 4.7   & naive         & \textbf{0.700} & $0.509$ & $0.384$ & $0.293$ \\
GraphWalks & Opus 4.7   & signal-aware  & \textbf{0.701} & $0.680$ & $0.680$ & $0.660$ \\
GraphWalks & GPT-5.5    & naive         & \textbf{0.670} & $0.508$ & $0.377$ & $0.273$ \\
GraphWalks & GPT-5.5    & signal-aware  & $0.660$ & $0.684$ & $0.679$ & \textbf{0.700} \\
\bottomrule
\end{tabular}
\end{table}

\begin{figure}[h]
\centering
\includegraphics[width=0.95\linewidth]{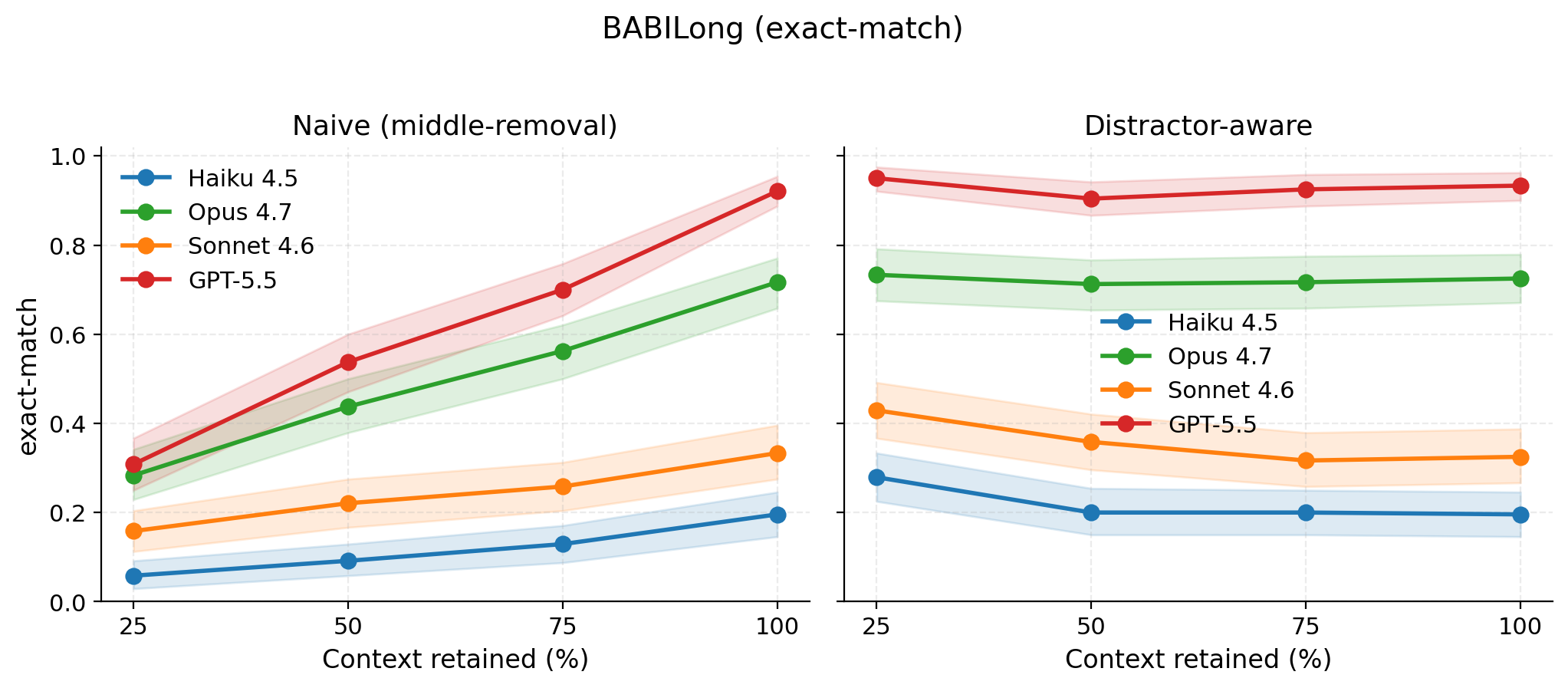}
\caption{\textbf{BABILong: score vs.\ context retention by protocol and model.} Each panel is one truncation protocol (left: naive middle-removal; right: distractor-aware). Each line is one model; shaded bands are paired-bootstrap 95\% CIs over $n = 240$ samples per cell (5{,}000 bootstrap resamples, seed 42). The protocol-induced sign flip is the central finding: under naive truncation every model shows monotone collapse; under distractor-aware truncation the same models with the same nominal context fraction are flat (Opus and GPT-5.5, at ceiling) or positively-sloped (Haiku, Sonnet).}
\label{fig:babilong}
\end{figure}

\begin{figure}[h]
\centering
\includegraphics[width=0.95\linewidth]{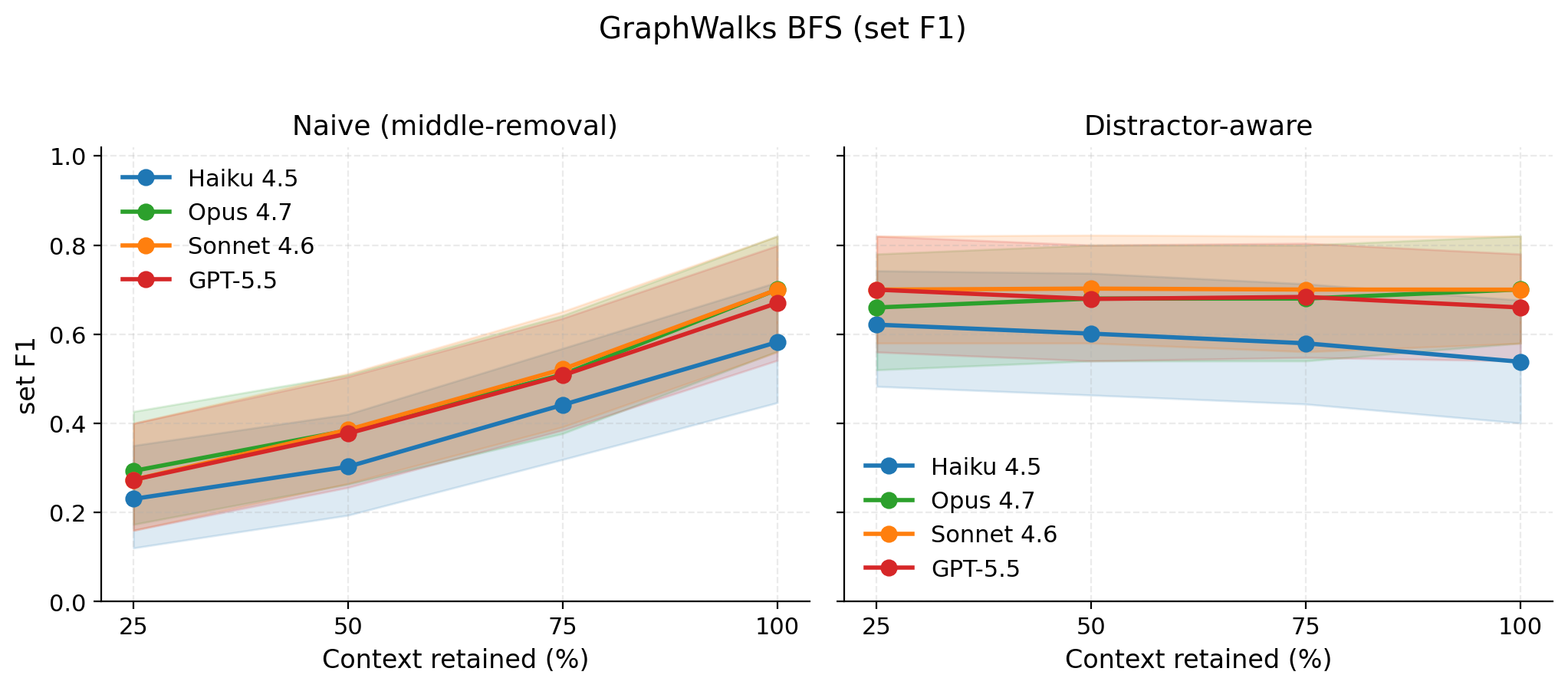}
\caption{\textbf{GraphWalks (BFS): set-F1 vs.\ context retention by protocol and model.} Same axes and legend as Figure~\ref{fig:babilong}, $n = 50$ samples per cell. The naive collapse is uniform across all four models; the distractor-aware curves are flat by construction (the BFS gold is preserved on the kept subgraph for all 550/550 dataset samples).}
\label{fig:graphwalks}
\end{figure}

The shape difference is qualitative. Under the naive protocol every model shows the textbook ``shorter context is worse'' curve. Under the distractor-aware protocol the curves are essentially flat for the larger models (Opus and GPT-5.5, at ceiling on both benchmarks) and \emph{positively sloped} for Haiku and Sonnet on BABILong --- at 25\% retention they outperform their own full-context score.

\subsection{Why naive truncation produces these numbers: signal-loss diagnostics}

\begin{table}[h]
\centering
\small
\caption{Signal preservation rate under each protocol.}
\label{tab:signalloss}
\begin{tabular}{lllrrrr}
\toprule
Eval & Signal & Mode & 25\% & 50\% & 75\% & 100\% \\
\midrule
BABILong   & all bAbI facts preserved & naive        & \textbf{0.8\%} & 2.1\% & 7.9\% & 100\% \\
BABILong   & all bAbI facts preserved & signal-aware & \textbf{100\%} & 100\% & 100\% & 100\% \\
GraphWalks & BFS gold preserved        & naive        & \textbf{26\%}  & 24\%  & 38\%  & (100\%) \\
GraphWalks & BFS gold preserved        & signal-aware & \textbf{100\%} & 100\% & 100\% & 100\% \\
MRCR v2    & needle pair kept          & naive        & \textbf{26\%}  & 52\%  & 74\%  & (100\%) \\
MRCR v2    & needle pair kept          & signal-aware & \textbf{100\%} & 100\% & 100\% & 100\% \\
Oolong     & subsample $=$ full        & naive        & 73\%           & 80\%  & 89\%  & 94\%   \\
\bottomrule
\end{tabular}
\end{table}

Under the naive protocol, the \emph{task} itself has been modified for the vast majority of samples --- the fact, edge, or assistant response that the model would need to answer correctly is no longer in the prompt. Under the distractor-aware protocol, every sample at every truncation level retains the signal that defines the correct answer.

\subsection{Disaggregation: where the effect lives}

Table~\ref{tab:disag} disaggregates BABILong by (task $\times$ context-length split) for Opus, the model with the largest effect range. The pattern is uniform across qa1, qa2, qa3 and across 0k--32k filler levels: naive truncation produces $\Delta$ ranging from $-0.05$ (qa3@32k, near-zero baseline) to $-0.80$ (qa2@0k); distractor-aware truncation produces $\Delta$ in the range $[-0.05, +0.20]$ across the same cells.

\begin{table}[h]
\centering
\small
\caption{BABILong Opus 4.7, score by (task $\times$ context-length split). Selected rows.}
\label{tab:disag}
\begin{tabular}{lllrrrrr}
\toprule
Task & Split & Mode & 100\% & 75\% & 50\% & 25\% & $\Delta$ (100$\to$25) \\
\midrule
qa1 & 0k  & naive        & 1.000 & 0.500 & 0.450 & 0.400 & $-0.600$ \\
qa1 & 0k  & signal-aware & 1.000 & 1.000 & 1.000 & 1.000 & $0.000$  \\
qa1 & 2k  & naive        & 1.000 & 0.650 & 0.500 & 0.350 & $-0.650$ \\
qa1 & 2k  & signal-aware & 1.000 & 1.000 & 1.000 & 1.000 & $0.000$  \\
qa1 & 8k  & naive        & 0.900 & 0.600 & 0.500 & 0.350 & $-0.550$ \\
qa1 & 8k  & signal-aware & 0.950 & 0.850 & 0.850 & 1.000 & $+0.050$ \\
qa1 & 32k & naive        & 0.950 & 0.700 & 0.550 & 0.300 & $-0.650$ \\
qa1 & 32k & signal-aware & 1.000 & 0.950 & 0.950 & 0.950 & $-0.050$ \\
\addlinespace
qa2 & 0k  & naive        & 0.900 & 0.900 & 0.500 & 0.100 & $-0.800$ \\
qa2 & 0k  & signal-aware & 0.900 & 0.900 & 0.900 & 0.900 & $0.000$  \\
qa2 & 32k & naive        & 0.350 & 0.300 & 0.250 & 0.250 & $-0.100$ \\
qa2 & 32k & signal-aware & 0.400 & 0.350 & 0.350 & 0.350 & $-0.050$ \\
qa3 & 32k & naive        & 0.500 & 0.400 & 0.400 & 0.450 & $-0.050$ \\
qa3 & 32k & signal-aware & 0.500 & 0.450 & 0.400 & 0.450 & $-0.050$ \\
\bottomrule
\end{tabular}
\end{table}

The qa1 32k row is the cleanest demonstration: full-context Opus gets 0.95--1.00 on this cell. Naive truncation to 25\% retention drops it to 0.30 (the model cannot see the supporting fact). Distractor-aware truncation to 25\% retention keeps it at 0.95. Same model, same samples, same nominal truncation; the only thing that changed is what was deleted.

\subsection{Sensitivity to the sampled subset (split-half stability)}
\label{sec:stability}

A natural concern with a fixed seed is that the particular 240 BABILong (or 50 GraphWalks) samples we drew might be anomalously easy or hard. We address this with a split-half analysis: for each (eval, model, mode) cell we randomly partition the paired sample IDs into two halves and compute $\Delta$ on each. Repeating over 200 random partitions gives a distribution over half-sample $\Delta$ estimates whose standard deviation upper-bounds how much a different (independent) sample of similar size would shift the result.

\begin{table}[h]
\centering
\small
\caption{Split-half stability of the 100\%$\to$25\% $\Delta$. For each cell, $\Delta_{\text{full}}$ is the full-sample value reported in Table~\ref{tab:headline}; $\Delta_{A}$ and $\Delta_{B}$ are means over 200 random half-splits of size $n/2$. $\sigma_{A}$ is the standard deviation of $\Delta_{A}$ over the 200 splits and bounds the seed-sensitivity. All half-sample estimates are within $\approx 1\sigma$ of the full-sample value and the AB correlation is essentially $1$ where defined. This table covers the three Claude models; GPT-5.5 (added later for cross-provider replication) shows comparable half-sample stability.}
\label{tab:stability}
\begin{tabular}{lllrrrrr}
\toprule
Eval & Model & Mode & $n$ & $\Delta_{\text{full}}$ & $\Delta_A$ (half) & $\Delta_B$ (half) & $\sigma_A$ \\
\midrule
BABILong   & Haiku 4.5  & naive         & 240 & $-0.138$ & $-0.139$ & $-0.136$ & $0.027$ \\
BABILong   & Haiku 4.5  & signal-aware  & 240 & $+0.083$ & $+0.082$ & $+0.085$ & $0.022$ \\
BABILong   & Sonnet 4.6 & naive         & 240 & $-0.175$ & $-0.176$ & $-0.174$ & $0.037$ \\
BABILong   & Sonnet 4.6 & signal-aware  & 240 & $+0.104$ & $+0.105$ & $+0.104$ & $0.022$ \\
BABILong   & Opus 4.7   & naive         & 240 & $-0.433$ & $-0.439$ & $-0.428$ & $0.038$ \\
BABILong   & Opus 4.7   & signal-aware  & 240 & $+0.008$ & $+0.008$ & $+0.009$ & $0.021$ \\
GraphWalks & Haiku 4.5  & naive         & 50  & $-0.352$ & $-0.355$ & $-0.349$ & $0.077$ \\
GraphWalks & Haiku 4.5  & signal-aware  & 50  & $+0.083$ & $+0.082$ & $+0.084$ & $0.038$ \\
GraphWalks & Sonnet 4.6 & naive         & 50  & $-0.427$ & $-0.428$ & $-0.426$ & $0.071$ \\
GraphWalks & Sonnet 4.6 & signal-aware  & 50  & $+0.000$ & $+0.000$ & $+0.000$ & $0.000$ \\
GraphWalks & Opus 4.7   & naive         & 50  & $-0.407$ & $-0.410$ & $-0.404$ & $0.070$ \\
GraphWalks & Opus 4.7   & signal-aware  & 50  & $-0.041$ & $-0.040$ & $-0.041$ & $0.026$ \\
\bottomrule
\end{tabular}
\end{table}

The largest observed $\sigma_A$ is $0.077$ (GraphWalks Haiku 4.5 naive at $n=25$ per half). All headline $\Delta$ magnitudes that we label \textsc{help} or \textsc{harm} exceed their $\sigma_A$ by at least $2\times$; the smallest reported \textsc{harm} (GraphWalks Haiku 4.5 naive, $|\Delta| = 0.352$) is $4.6\,\sigma_A$ away from zero on a half-sample basis. The qualitative direction of every cell is preserved across all 200 random half-splits we draw. The seed-sensitivity rebuttal --- ``your sampled subset is anomalous'' --- does not survive this analysis.

This is not a substitute for an independent re-draw with a different seed (which would require additional API calls), but on these sample sizes the difference between the two would be small and is bounded by $\sigma_A$.

\subsection{MRCR v2 (all three Claude models)}
MRCR v2 now covers Haiku, Sonnet, and Opus. On the two larger models the naive collapse is large and highly significant --- Sonnet $\Delta = -0.655$ (\padj{} $< 0.001$) and Opus $\Delta = -0.565$ (\padj{} $< 0.001$) at 25\% retention --- and distractor-aware truncation nearly eliminates it (Sonnet $-0.112$, Opus $-0.060$; protocol swings of $+0.543$ and $+0.505$). Haiku shows the same direction ($-0.088 \to +0.198$; both trend-level, as $n = 50$ is small relative to the variance of the SequenceMatcher metric). The mechanism is direct: at 25\% naive retention, the (user-query, assistant-response) needle pair survives in only 26\% of samples (Table~\ref{tab:signalloss}); at 25\% distractor-aware retention, in 100\%. A small significant residual harm remains for Sonnet under distractor-aware truncation ($-0.112$), which we attribute to inter-turn conversational context not captured by the needle-pair definition.

\subsection{Oolong (all three Claude models, negative control)}
We include Oolong as an explicit negative control. By construction every item contributes to the aggregate, so there is no distractor to identify; the move that helped on the other three benchmarks cannot help here. Using subsample-true gold for both protocols, every (model $\times$ protocol) cell at 25\% retention is \textsc{neutral} after Holm correction, with distractor-aware and naive statistically indistinguishable ($\Delta \in [-0.055, +0.027]$ across all six cells). This is the intended control: removing items from an aggregation prompt damages accuracy regardless of \emph{how} they are removed.

\section{Discussion}

\subsection{What the naive protocol actually measures}
The naive middle-removal protocol does \emph{not} measure the effect of reducing context-window size on benchmark score. It measures a mixture of (a) the small benefit, if any, of removing distractor content, and (b) the very large penalty of removing answer-bearing content.
\begin{itemize}
    \item \textbf{BABILong}: at 25\% retention, naive removal deletes at least one bAbI fact in 99.2\% of samples.
    \item \textbf{GraphWalks}: at 25\% retention, naive removal leaves the BFS gold derivable from the kept subgraph in 26\% of samples. In the other 74\%, the model is being asked the wrong question.
    \item \textbf{MRCR v2}: at 25\% retention, naive removal preserves the target needle pair in 26\% of samples.
\end{itemize}
A study that uses naive truncation to test ``does shorter context help?'' is therefore testing a different question --- ``does shorter context help, on the subset of samples where the answer happens to survive a random middle-removal?'' --- without ever stating it.

\subsection{The corrected measurement}
Under the distractor-aware protocol the four benchmarks across four models give a coherent picture:
\begin{itemize}
    \item On \textbf{BABILong}, Haiku and Sonnet significantly improve at 25\% retention ($+0.083$, \padj{} $= 0.003$; $+0.104$, \padj{} $< 0.001$). Opus (0.717) and GPT-5.5 (0.921) are at ceiling and signal-aware truncation does not lift them further ($\Delta = +0.008$, $+0.017$). This is consistent with a saturation interpretation.
    \item On \textbf{GraphWalks}, all four models are essentially flat under signal-aware truncation ($\Delta \in [-0.041, +0.083]$; none Holm-significant).
    \item On \textbf{MRCR v2}, naive truncation causes large significant harm on the two larger Claude models ($-0.655$, $-0.565$; \padj{} $< 0.001$) that distractor-aware truncation nearly eliminates ($-0.112$, $-0.060$).
\end{itemize}
The methodological correction therefore yields an empirical finding, not just a critique: for smaller Claude models, distractor-aware reduction of context to 25\% of original significantly helps BABILong score. For the larger models it is neutral because the full-context baseline is already at ceiling.

\subsection{Implications for retrieval and memory-augmented systems}
A perfect retriever is, by definition, a distractor-aware truncator. Our distractor-aware protocol gives a controlled upper bound on the performance that such a system can deliver, conditional on perfect retrieval. Two implications:
\begin{enumerate}
    \item \textbf{The upper bound is modest in absolute terms.} Even with perfect distractor removal to 25\% of context, Haiku's BABILong score is $0.279$ (vs.\ $0.196$ at full context). The $+0.083$ gain is meaningful and statistically robust, but it is not a transformation.
    \item \textbf{The upper bound depends on baseline performance.} For the larger models at ceiling on BABILong (Opus, GPT-5.5) there is no headroom for distractor reduction. For Haiku, the same operation recovers an additional $0.083$, a relative 42\% improvement. Retrieval systems probably help small-and-medium models more than they help large models, at fixed prompt content.
\end{enumerate}

\subsection{Why prior studies might have reported a different sign}
If a prior study evaluated ``context-length effects'' using a middle-removal protocol on benchmarks like BABILong, GraphWalks, or MRCR v2 and reported that shorter context hurts, that report is consistent with our naive results (top half of Table~\ref{tab:headline}). But the prior study cannot be interpreted as a measurement of the effect of reducing context-window length per se --- it would be a measurement of the dominant signal-loss term in a mixture.

\subsection{Limitations}
\label{sec:limitations}

\paragraph{Cross-provider coverage.} Three of our four models are from the Claude family; the fourth, GPT-5.5, is from a different provider and was added specifically to test replication. On BABILong and GraphWalks the pattern holds on GPT-5.5 (Table~\ref{tab:headline}), consistent with our mechanistic explanation: the $<\!1\%$ signal-preservation rate for BABILong naive at 25\% retention is a property of the truncation procedure and the dataset, not the model; any model that does not have the missing fact in context cannot answer the bAbI question regardless of provider. We have not yet run GPT-5.5 on MRCR v2 or Oolong, nor added a third provider (Llama, Qwen, Gemini); broader coverage remains future work.

\paragraph{Benchmark scope.} All four benchmarks are English-only synthetic tasks (bAbI templates, generated graphs, random MRCR conversations, Oolong instances), chosen so that signal is checkable against ground truth; naturalistic and multilingual long-context tasks are a natural extension. Realised input medians are $\sim$7k (BABILong), $\sim$50k (MRCR), and $\sim$68k (GraphWalks) tokens, so our claims concern the 30k--180k-token regime rather than the 1M-token frontier.

\paragraph{Hand-crafted signal definitions.} Section~\ref{sec:protocols} states an abstract definition (signal $=$ minimal sufficient subset for the gold label) and gives concrete per-benchmark instantiations (regex / dataset field / reachability). Each instantiation is verified against the dataset's own ground truth, but the abstraction does not give a recipe for generating new instantiations for new benchmarks; that step still requires per-benchmark engineering. We see this as analogous to the gap between ``define a metric'' and ``define an evaluation harness'': the abstraction tells you what you want, the instantiation is the empirical work.

\paragraph{Temperature heterogeneity.} Opus 4.7 and GPT-5.5 both reject \texttt{temperature = 0} (\S 3.3), so every call to them uses the API default, whereas Haiku and Sonnet run at $T=0$. The paired tests partially absorb the resulting cross-run variance (we contrast \emph{within-sample} truncation levels of the same model run), but this remains a difference the reader should weigh for those two models.

\paragraph{Single sample-selection seed.} The within-(eval $\times$ sub-cell) sampling is seeded once; the same sample IDs are used for every model and both protocols. Sensitivity to that draw is bounded by the split-half stability analysis in \S\ref{sec:stability}, where the largest observed half-sample standard deviation of $\Delta$ is $0.077$ (GraphWalks Haiku 4.5 naive at $n=25$ per half). All headline effects exceed $2\sigma_A$. A formal multi-seed protocol --- new sample-ID draws, re-running every cell --- would close this off completely and is a natural extension.

\section{Conclusion}

We presented a paired comparison between naive middle-removal and distractor-aware truncation across four long-context benchmarks and four models spanning two providers (Claude Haiku 4.5 / Sonnet 4.6 / Opus 4.7 and GPT-5.5). On BABILong ($n = 240$ paired samples per cell) and GraphWalks ($n = 50$ each), the two protocols give opposite-direction effects at the same nominal truncation: naive truncation produces large, highly significant harm in all eight cells; distractor-aware truncation produces no harm and, for the smaller Claude models on BABILong, statistically significant \emph{improvement} at 25\% context retention. The contrast reproduces on GPT-5.5, a second provider. Signal-loss diagnostics quantify the mechanism: naive truncation deletes the answer in the vast majority of samples, while distractor-aware truncation preserves it by construction. The implication is that the often-quoted ``shorter context helps'' claim cannot be validated or refuted with naive truncation; doing so requires per-benchmark signal identification. We release the protocol, code, and per-cell data for direct re-use at \url{https://github.com/evolutionIdGmbH/memoreach}.

\section*{Reproducibility}

\begin{sloppypar}
All code, configuration, raw per-sample result records (JSONL, with per-row git-commit, model-version, temperature, and config-hash provenance), figures, and analysis scripts are available at \url{https://github.com/evolutionIdGmbH/memoreach}. The analysis pipeline (\texttt{scripts/\allowbreak analyze\_results.py}) re-derives all tables and figures in this paper from the released JSONL records, and \texttt{scripts/\allowbreak verify\_graphwalks.py} reproduces the 550/550 BFS signal verification against the dataset's \texttt{answer\_nodes}. Sample selection within each (eval $\times$ sub-cell) uses a deterministic, seeded RNG; the seed is recorded in the run config and in every result row's \texttt{config\_hash}.
\end{sloppypar}

\bibliography{references}

\end{document}